\documentclass[conference]{IEEEtran}
\IEEEoverridecommandlockouts
\usepackage{cite}
\usepackage{amsmath,amssymb,amsfonts}
\usepackage{algorithmic}
\usepackage{graphicx}
\usepackage{textcomp}
\usepackage{xcolor}

\usepackage[ruled, vlined, linesnumbered]{algorithm2e}
\usepackage[colorlinks, linkcolor=red, citecolor=blue]{hyperref}
\usepackage{multirow}
\usepackage{color}
\usepackage{bbding}
\usepackage{listings}
\def\ourmethod{{STCMOT}\xspace} 
\def\BibTeX{{\rm B\kern-.05em{\sc i\kern-.025em b}\kern-.08em
    T\kern-.1667em\lower.7ex\hbox{E}\kern-.125emX}}
\begin{document}

\title{STCMOT: Spatio-Temporal Cohesion Learning for UAV-Based Multiple Object Tracking\\
\thanks{* Corresponding author}
}

\author{
\IEEEauthorblockN{1\textsuperscript{st} Jianbo Ma}
\IEEEauthorblockA{
\textit{National Key Laboratory of Optical Field} \\
\textit{Manipulation Science and Technology}\\
\textit{Key Laboratory of Optical Engineering}\\
\textit{Institute of Optics and Electronics}\\
\textit{University of Chinese Academy of Sciences}\\
Chengdu, China\\
majianbo22@mails.ucas.ac.cn}
\and
\IEEEauthorblockN{2\textsuperscript{nd} Chuanming Tang}
\IEEEauthorblockA{
\textit{Institute of Optics and Electronics}\\
\textit{University of Chinese Academy of Sciences}\\
Chengdu, China\\
tangchuanming19@mails.ucas.ac.cn}
\and
\IEEEauthorblockN{3\textsuperscript{rd} Fei Wu}
\IEEEauthorblockA{
\textit{Pattern Recognition Lab}  \\
\textit{Department of Computer Science} \\
\textit{Friedrich-Alexander-Universität} \\
\textit{Erlangen-Nürnberg} \\
Erlangen, Germany \\
river.wu@fau.de}
\and
\IEEEauthorblockN{4\textsuperscript{th} Can Zhao}
\IEEEauthorblockA{
\textit{Institute of Optics and Electronics}\\
\textit{University of Chinese Academy of Sciences}\\
Chengdu, China\\
zhaocan22@mails.ucas.ac.cn}
\and
\IEEEauthorblockN{5\textsuperscript{th} Jianlin Zhang*}
\IEEEauthorblockA{
\textit{Key Laboratory of Optical Engineering}\\
\textit{Institute of Optics and Electronics}\\
Chengdu, China\\
jlin@ioe.ac.cn}
\and
\IEEEauthorblockN{6\textsuperscript{th} Zhiyong Xu}
\IEEEauthorblockA{
\textit{Key Laboratory of Optical Engineering}\\
\textit{Institute of Optics and Electronics}\\
Chengdu, China\\
xuzhiyong@ioe.ac.cn}
}

\maketitle

\begin{abstract}
Multiple object tracking (MOT) in Unmanned Aerial Vehicle (UAV) videos is important for diverse applications in computer vision. Current MOT trackers rely on accurate object detection results and precise matching of target re-identification (ReID). These methods focus on optimizing target spatial attributes while overlooking temporal cues in modelling object relationships, especially for challenging tracking conditions such as object deformation and blurring, etc.
To address the above-mentioned issues, we propose a novel \textbf{S}patio-\textbf{T}emporal \textbf{C}ohesion \textbf{M}ultiple \textbf{O}bject \textbf{T}racking framework (\ourmethod), which utilizes historical embedding features to model the representation of ReID and detection features in a sequential order. Concretely, a temporal embedding boosting module is introduced to enhance the discriminability of individual embedding based on adjacent frame cooperation. While the trajectory embedding is then propagated by a temporal detection refinement module to mine salient target locations in the temporal field. Extensive experiments on the VisDrone2019 and UAVDT datasets demonstrate our \ourmethod sets a new state-of-the-art performance in MOTA and IDF1 metrics.
The source codes are released at \href{https://github.com/ydhcg-BoBo/STCMOT}{https://github.com/ydhcg-BoBo/STCMOT}.
\end{abstract}


\section{Introduction}
The goal of multiple object tracking (MOT) is to recognize various designated target categories in video sequences.
In recent years, the MOT algorithms for unmanned aerial vehicle (UAV) videos have captured the interest of researchers. 
Its primary task lies in effectively tracking multiple objects within UAV views.
% from the perspective of UAV. 
However, with the distance between the objects and the UAV camera, objects are typically small and prone to blurring. This results in many challenging attributes for the UAV-based MOT task~\cite{uav-survey}.
Unlike stationary shooting, the dynamic motion of the UAV introduces continuous object position changes and multi-dimensional scaling issues related to the ground objects. 
As a result, the performance of UAV tracking is easily impaired by interference from similar distractors and occlusions.

\begin{figure}[tb]
    \centering
    \includegraphics[scale=0.35]{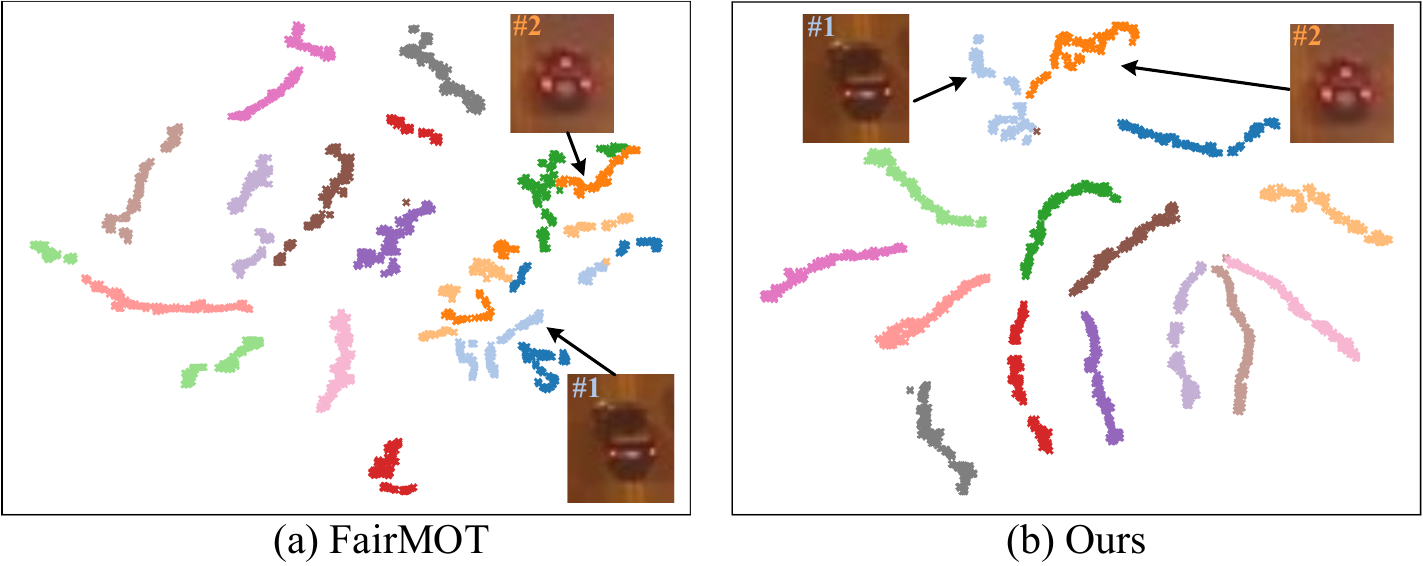}
    \caption{T-SEN projection of ReID embeddings for the 15 tracked targets in the UAVDT-M1007 video.  \ourmethod shows a more discriminative embedding representation compared with the baseline FairMOT, even in the cases where targets \#1 and \#2 have similar appearances during nighttime.}
    \label{tsne}
    \vspace{-0.4cm}
\end{figure}

To address these challenges, previous works~\cite{MOTDT, deepsort} mainly utilize the two-stage tracking framework to enhance tracking capability.
They first introduce a detection model to identify the objects' positions. Subsequently, the motion information and re-identification (ReID) embeddings of object candidates are established and connected to complete the matching.
Despite the remarkable tracking performance from these two-stage trackers~\cite{MOTDT, deepsort, zhang2022bytetrack, cao2023observation}, using the respective network for object detection and embedding extraction results in considerable storage cost and resource consumption~\cite{Aharon2022BoTSORTRA}.
To this end, the one-shot tracking algorithms~\cite{wang2020JDE, zhang2021fairmot, FDTrack} integrate the detection branch and the ReID branch into a unified framework for balancing tracking performance and speed. 
For example, FairMOT~\cite{zhang2021fairmot} merges an additional ReID branch into the CenterNet~\cite{zhou2019objects}. 
It takes a single-frame image as input and simultaneously produces the detected target and corresponding embedding features.
This streamlined tracking framework can simplify the tracking process and thus is more suitable for real-world object tracking in UAV videos. 
\begin{figure*}[htbp]
    \centerline{\includegraphics[scale=1.10]{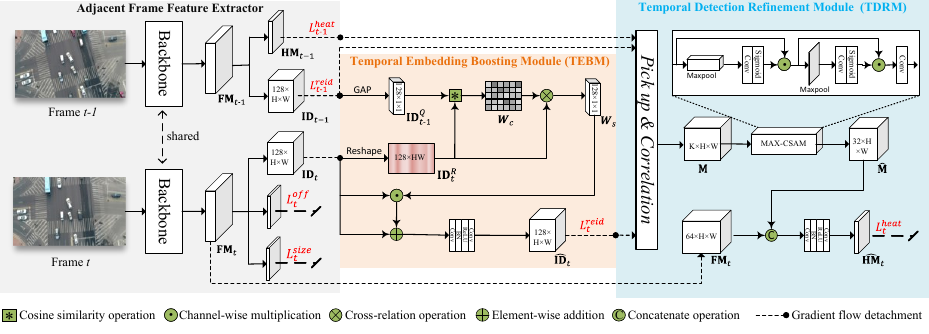}}
    \caption{Architecture details of \ourmethod. It consists of three components: a Frame Feature Extractor, a Temporal Embedding Boosting Module, and a Temporal Detection Refinement Module.}
    \label{pipeline}
    \vspace{-0.4cm}
\end{figure*}

Additionally, propelled by the advancement of deep convolutional networks, some works~\cite{fpuav,GCEVT,yu2022relationtrack,liang2022cstrack} use enhanced deep features in the one-shot tracking framework and achieve higher performance.
To obtain pixel-wise relations at multiple feature levels, Wu et al.~\cite{fpuav} firstly design a pyramid transformer encoder for the detection branch, then a channel-wise transformer encoder which equipped with the pyramid fusion network~\cite{GCEVT} is used to enhance the ReID branch.
At the same time, Yu et al.~\cite{yu2022relationtrack} propose a guided transformer encoder to emphasize global semantic information. 
Liang et al.~\cite{liang2022cstrack} introduce a scale-aware attention network to alleviate object semantic misalignments.
In addition to enhancing the feature representation from global and local spatial perspectives, Liu et al.~\cite{liu2022uavmot} model the temporal relationship of ID features and propose an ID feature update module to adapt to appearance changes in various UAV views.
However, the multiple object spatio-temporal cues from the detection branch have not yet been seriously considered in these works, which inspires us to construct a spatio-temporal interaction model between detection and ReID branches to handle complex tracking scenarios. E.g., object deformation and blurring.

In this paper, we propose a unified spatio-temporal cohesion learning network for MOT in UAV-captured videos, termed \ourmethod.
\ourmethod extends the salient ReID embeddings from the historical frame to the current frame, optimizing and enhancing both detection and ReID branches. 
In particular, we first introduce a temporal embedding boosting module. It couples the ReID feature map from adjacent frames and generates a channel-wise descriptor to highlight the discriminability of individual embedding.
Secondly, a temporal detection refinement module is built upon the inherent continuity of individual embedding in video sequences. This module regards each previous target embedding as a query, facilitating information interactions with the current ReID feature map during one-shot network training.

In this way, the object spatio-temporal coupling strategy can optimize the detection and ReID branches at the same time, thus strengthening the network association capability and reducing tracking failures.
The experimental results demonstrate that the proposed method surpasses the current state-of-the-art trackers in MOTA and IDF1 metrics.
Overall, the main contributions of this work can be summarized in the following three aspects:
\begin{itemize}
  \item We emphasize the importance of introducing the ReID feature map from successive frames in MOT. A temporal embedding boosting module (named TEBM) is proposed to improve the specificity of individual embedding at the feature channel level.
  \item We propose a temporal detection refinement module named TDRM. It is designed to leverage the object moving continuity and trajectory embedding for refining fine-grained object position and enhancing detection performance. 
  \item With the proposed spatio-temporal semantic embedding cohesion strategy, 
  % \ourmethod achieves an IDF1 of 52.0\% and a MOTA of 41.2\% on the VisDrone2019 dataset, and an IDF1 of 69.3\% and a MOTA of 48.9\% on the UAVDT dataset, respectively.
  \ourmethod achieves new state-of-the-art performance on both VisDrone2019 and UAVDT benchmarks. 
\end{itemize}

\section{Methodology}

\subsection{Overall Framework}
As shown in Fig.~\ref{pipeline}, \ourmethod can be divided into three parts: an adjacent frame feature extractor, a temporal embedding boosting module (TEBM), and a temporal detection refinement module (TDRM). 
The overall process can be described as follows:
First, the UAV platform collects a video sequence and feeds two adjacent frames, e.g., frame $t-1$ and frame $t$, into our feature extractor.
Then, our ReID module augments the ReID embeddings as $\hat{\textbf{ID}}_{t}\in\mathbb{R}{^{{{128}} \times {{H}} \times {{W}}}}$, our detection module refines the detection heatmap as $\hat{\textbf{HM}}_{t}\in\mathbb{R}{^{{{C}} \times {{H}} \times {{W}}}}$.
Here, $C$ denotes the target category. $H, W$ are the height and weight of the corresponding feature map.
Specifically, TEBM enhances the robustness and saliency of individual embedding by combining the ReID feature map from adjacent frames.
Based on the embedding inherent continuity of the tracked trajectories throughout the entire video sequence, TDRM is proposed to restore and enhance the location of targets in the detection stage.

\noindent
\textbf{Feature Extractor.} As shown in Fig.~\ref{pipeline}, our framework adopts a weight-shared backbone (DLA-34~\cite{zhou2019objects}) to get the original feature map of the previous frame $\textbf{FM}_{t-1}\in\mathbb{R}{^{{{64}} \times {{H}} \times {{W}}}}$ and current frame $\textbf{FM}_{t}\in\mathbb{R}{^{{{64}} \times {{H}} \times {{W}}}}$.
The original feature map $\textbf{FM}_{t-1}$ is then passed through a separate 3 $\times$ 3 convolution layer, a ReLU layer and a 1 $\times$ 1 convolution layer to obtain the detection heatmap $\textbf{HM}_{t-1}\in\mathbb{R}{^{{{C}} \times {{H}} \times {{W}}}}$ and ReID feature map $\textbf{ID}_{t-1}\in\mathbb{R}{^{{{128}} \times {{H}} \times {{W}}}}$. 
Also, we obtain initial results from $\textbf{FM}_{t}$, which include ReID feature map $\textbf{ID}_{t}\in\mathbb{R}{^{{{128}} \times {{H}} \times {{W}}}}$, box offset of $\mathbb{R}{^{{{2}} \times {{H}} \times {{W}}}}$, and box size of $\mathbb{R}{^{{{2}} \times {{H}} \times {{W}}}}$.

% For each head, the original feature map is then passed through a separate 3 $\times$ 3 convolution layer, ReLU and another 1 $\times$ 1 convolution layer to obtain the detection heatmap $\textbf{HM}_{t-1}\in\mathbb{R}{^{{{C}} \times {{H}} \times {{W}}}}$ and ReID embeddings $\textbf{ID}_{t-1}\in\mathbb{R}{^{{{128}} \times {{H}} \times {{W}}}}$ from $\textbf{FM}_{t-1}$. 
% We also obtain initial results from $\textbf{FM}_{t}$, which include ReID embeddings $\textbf{ID}_{t}\in\mathbb{R}{^{{{128}} \times {{H}} \times {{W}}}}$, box offset of $\mathbb{R}{^{{{2}} \times {{H}} \times {{W}}}}$, and box size of $\mathbb{R}{^{{{2}} \times {{H}} \times {{W}}}}$.

\subsection{Temporal Embedding Boosting Module}
In the MOT system, ReID embeddings play a crucial role in online association.
It generally sustains a continuous trajectory through similarity matching in adjacent frames.
Nevertheless, UAV motion often introduces variations in object scale and position, which simultaneously weakens the ReID learning process and online association.
To this end, we propose TEBM to enhance the ReID learning process from temporal cues.
% which is harmful to the ReID learning process and online association at the same time.
% Our introduced TEBM module improves the ReID process with temporal cues. 

As shown in the middle part of Fig.~\ref{pipeline}, TEBM takes the ReID feature map $\textbf{ID}_{t-1}$ and $\textbf{ID}_t$ as inputs.
To obtain the statistical distribution information of the whole feature map, global average pooling (GAP) is firstly introduced to project the ReID information of the previous frame \(\textbf{ID}_{t-1}\) as a 128-dimensional operator $\textbf{ID}_{t-1}^{Q}$.
% the adjacent frames as input.
% First, global average pooling (GAP) is introduced to inject the ID information in the previous frame \(\textbf{ID}_{t-1}\) as a 128-dimensional operator $\textbf{ID}_{t-1}^{Q}$. It can get the average distribution of the whole map.
Then we reshape the current frame \(\textbf{ID}_t \in \mathbb{R}^{128 \times H \times W}\) into  \(\textbf{ID}_t^R \in \mathbb{R}^{128 \times HW}\).
Lastly, we calculate the cosine similarity between the query operator \(\textbf{ID}_{t-1}^{Q}\) and the key $\textbf{ID}_t^R$, which can be expressed as:
% Lastly, with the input query operator \(\textbf{ID}_{t-1}^{Q}\) and the key $\textbf{ID}_t^R$, we calculate the cosine similarity operation between them, which can be expressed as:
\begin{equation}
    % \textit{\textbf{W}}_c = \frac{\textbf{ID}_{t-1}^{Q}  \cdot \textbf{ID}_t^R }{\|\textbf{ID}_{t-1}^{Q}\| \cdot \|\textbf{ID}_t^R \|}
    \textit{\textbf{W}}_c = \textbf{Cos}(\textbf{ID}_{t-1}^{Q}, \textbf{ID}_t^R)= \frac{\textbf{ID}_{t-1}^{Q}  \cdot \textbf{ID}_t^R }{\|\textbf{ID}_{t-1}^{Q}\| \cdot \|\textbf{ID}_t^R \|}
\end{equation}
where $\textit{\textbf{W}}_c\in \mathbb{R}^{H \times W}$ is the salient attention matrix that measures the positional relationship of each instance embedding in two frames. 

Following this, a cross-relation operation is employed to generate channel-level weighted descriptor \(\textit{\textbf{W}}_s\) for the ReID feature map. Its calculation is as follows:
\begin{equation}
    \textit{\textbf{W}}_s = LN(Conv(\textbf{ID}_t^R \cdot \textit{\textbf{W}}_c))
\end{equation}
where $Conv$ denotes 1 $\times$ 1 convolution layer while $LN$ denotes LayerNorm layer. Finally, we employ \(\textit{\textbf{W}}_s\) to reweight the current ReID embeddings, as depicted in Eq.~\ref{boosting_id}. 
This approach enables the model to effectively concentrate on specific information across different channels, thus positively contributing to the extraction of discriminative embedding.
\begin{equation}\label{boosting_id}
    \hat{\textbf{ID}}_{t} = \psi(\textbf{ID}_t \odot \textit{\textbf{W}}_s \oplus \textbf{ID}_t)
\end{equation}

Here, \(\odot\) represents channel-wise multiplication, \(\oplus\) represents element-wise addition, and \(\psi\) denotes a Conv2d-BatchNorm-ReLU-Conv2d block used to obtain the final boosted ReID embeddings $\hat{\textbf{ID}}_t$.
\begin{algorithm}[t]
\small
\caption{\small Pseudo-code of Pick up \& Correlation.}
\label{alg:puc}
\definecolor{codeblue}{rgb}{0.25,0.5,0.5}
\definecolor{codekw}{rgb}{0.85, 0.18, 0.50}
\lstset{
  backgroundcolor=\color{white},
  basicstyle=\fontsize{8pt}{8pt}\ttfamily\selectfont,
  columns=fullflexible,
  breaklines=true,
  captionpos=b,
  commentstyle=\fontsize{8pt}{8pt}\color{codeblue},
  keywordstyle=\fontsize{8pt}{8pt}\color{codekw},
  escapechar={|}, 
}
\begin{lstlisting}[language=python,mathescape=true,escapeinside={(*@}{@*)}]
import torch
import torch.nn.function as F

def Pick_and_Corr($\textbf{HM}_{t-1}$, $\textbf{ID}_{t-1}$, $\hat{\textbf{ID}}_{t}$, $\textbf{K}=100$):
    $\textbf{HM}_{t-1}^{max}$ = F.maxpool_2d($\textbf{HM}_{t-1}$)
    $\textbf{HM}_{t-1}$ = $\textbf{HM}_{t-1}$ * ($\textbf{HM}_{t-1}$ == $\textbf{HM}_{t-1}^{max}$)
    $\textbf{HM}_{t-1}$ = $\textbf{HM}_{t-1}$.view(-1, C*H*W)
    topK_scores, topK_inds = torch.topk($\textbf{HM}_{t-1}$, $\textbf{K}$)
    $\textbf{ID}_{K} = \textbf{ID}_{t-1}(\text{topK\_inds}<K)$
    $\textbf{M}$ = torch.matmul($\textbf{ID}_{K}$, $\hat{\textbf{ID}}_{t}$)
    return $\textbf{M}$
\end{lstlisting}
\end{algorithm}

\subsection{Temporal Detection Refinement Module}
Unlike single-frame detection task, the MOT system processes sequential video frames, making the temporal context information as important as spatial information.
Therefore, mining temporal trajectory information and integrating spatio-temporal features can effectively enhance MOT detection performance.
As shown in the right part of Fig.~\ref{pipeline}, our TDRM is designed to fully interact with the temporal information between two adjacent frames for fine-grained detection.

% Unlike single-frame detection tasks, the MOT system processes video sequences frame by frame, making the temporal information contribute much to the object tracking except for the general spatial.
% Therefore, mining temporal trajectory information and integrating temporal-spatial features can effectively enhance MOT detection performance.
% As shown in the right of Fig.~\ref{pipeline}, in our detection branch, TDRM fully employs the temporal information between two frames for fine-grained detection.

The top K response probability map $\textbf{M}\in \mathbb{R}^{K \times H \times W}$ which generated from the $\textit{Pick up \& Correlation}$ strategy is detailed in Algorithm~\ref{alg:puc}.
Specifically, we first perform a $3\times3$ max pooling operation on the previous heatmap $\textbf{HM}_{t-1}$ and then select the top K locations of high-response points on $\textbf{HM}_{t-1}$. 
Afterwards, the top K embeddings $\textbf{ID}_{K} \in \mathbb{R}^{K \times 128}$ are extracted from $\textbf{ID}_{t-1}$ based on their location constraint.
Lastly, the probability map \textbf{M} is computed using the \textit{Correlation} operator between $\textbf{ID}_{K}$ and $\hat{\textbf{ID}}_{t}$, which represents the likelihood that the trajectory from the previous frame is existent in the current frame.

% At the beginning, we gain the Top K similarity maps $\textbf{M}\in \mathbb{R}^{K \times H \times W}$ by the $\textit{Pick up \& Correlation}$ strategy. 
% The whole process is shown in Algorithm~\ref{alg:puc}.
% Firstly, we perform a $3\times3$ max pooling operation on the previous heatmap $\textbf{HM}_{t-1}$ and select the top K locations of high-response points on $\textbf{HM}_{t-1}$. Then, the Top-K embeddings $\textbf{ID}_{K} \in \mathbb{R}^{K \times 128}$ is selected from the $\textbf{ID}_{t-1}$ based on their location. 
% The probability \textbf{M} is then obtained by using the \textit{Correlation} operator between $\textbf{ID}_{K}$ and $\hat{\textbf{ID}}_{t}$, which represents the likelihood that the trajectory from the previous frame exists in the current frame.

With the probability map \textbf{M}, a maximum channel spatial attention module (MAX-CSAM) is used to compress the channel dimension from K to 32 to generate the aggregated similarity map $\hat{\textbf{M}}$, formatted as:
\begin{equation}
   \hat{\textbf{M}} = \sigma(f_c(\textbf{M}_c^{max}))  \odot \sigma(f_s(\textbf{M}_s^{max})) \odot \textbf{M}
\end{equation}
where $\textbf{M}_c^{max}$ and $\textbf{M}_s^{max}$ are acquired through max pooling on the channel dimension and spatial dimension, respectively.
$f_c$ denotes a 1D convolution layer, $f_s$ denotes a $7 \times 7$ convolution layer and $\sigma$ is Sigmoid operator.
Then, we concatenate the original feature map $\textbf{FM}_t$ and $\hat{\textbf{M}}$, and pass them through a Conv2d-BatchNorm-ReLU-Conv2d block to obtain the refined detection heatmap $\hat{\textbf{HM}}_t$.
It can be expressed as:
\begin{equation}
   \hat{\textbf{HM}}_t = \psi ([\textbf{FM}_t, \hat{\textbf{M}}])
\end{equation}
where $[ , ]$ denotes the concatenate operation. 
The final detection heatmap $\hat{\textbf{HM}}_t$ fuses prior trajectory information to increase the probability of target appearance and focuses on reducing missed detections.

\begin{table*}[ht]
    \centering
    \caption{Quantitative comparison with the existing state-of-the-art methods on VisDrone2019 and UAVDT test sets.}
    \resizebox{1.0\linewidth}{!}{
    \begin{tabular}{c|c|c|cccccccc}
    \hline
    Dataset &Method  &Pub\&Year &IDF1$\uparrow(\%)$ & MOTA$\uparrow(\%)$ &MT$\uparrow$ &ML$\downarrow$  & FP$\downarrow$ & FN$\downarrow$   & IDS$\downarrow$ \\
    \hline
    \multirow[m]{8}{*}{VisDrone2019}
    & SORT~\cite{sort} &ICIP2016 &38.0 &14.0 &506 &545  &80845 &112954 &3629\\
    & IOUT~\cite{iout} &AVSS2017 &38.9 &28.1 &467 &670  &36158 &126549 &2393 \\
    & MOTR~\cite{zeng2022motr} &ECCV2022 &41.4 &22.8 &272 &825  &28407 &147937 &\textbf{959}  \\
    & TrackFormer~\cite{meinhardt2022trackformer} &CVPR2022 &30.5 &25.0 &385 &770   &25856 &141526 &4840 \\
    & ByteTrack~\cite{zhang2022bytetrack} &ECCV2022 &37.0 &35.7 &- &-  &21434 &124042 &2168\\
    & UAVMOT~\cite{liu2022uavmot} &CVPR2022 &51.0 &36.1 &520 &574   &27983 &115925 &2775\\
    & OCSORT~\cite{cao2023observation} &CVPR2023 &50.4 &39.6 &- &-  &\textbf{14631} &123513  &986\\
    & \ourmethod &Ours &\textbf{52.0} &\textbf{41.2} &\textbf{667} &\textbf{453}  &36428 &\textbf{94445} &3984\\
    \hline
    \multirow[m]{7}{*}{UAVDT}
    & SORT~\cite{sort} &ICIP2016 &43.7 &39.0 &484 &400   &33037 &172628 &2350 \\
    & IOUT~\cite{iout} &AVSS2017 &23.7 &36.6 &534 &357 &42245 &163881 &9938 \\
    & DSORT~\cite{deepsort}&ICIP2017 &58.2 &40.7 &595 &358 &44868 &155290 &2061 \\
    & ByteTrack~\cite{zhang2022bytetrack} &ECCV2022 &59.1 &41.6 &- &-  &\textbf{28819} &189197 &296\\
    & UAVMOT~\cite{liu2022uavmot} &CVPR2022 &67.3&46.4 &624 &221  &66352 &115940 &456 \\
    & OCSORT~\cite{cao2023observation} &CVPR2023 &64.9 &47.5 &- &-  &47681 &148378 &\textbf{288}\\
    & \ourmethod &Ours &\textbf{69.8} &\textbf{49.2} &\textbf{664} &\textbf{203} &72901 &\textbf{99547} &665\\
    \hline
    \end{tabular}
    }
    \label{test}
    \vspace{-0.4cm}
\end{table*}

\subsection{Objective Function}
With the consecutive frame pipeline, joint loss functions are employed to supervise the model training process.
As depicted in Fig.~\ref{pipeline}, the previous frame $t-1$ is supervised with ReID loss $L_{t-1}^{reid}$ and heatmap loss $L_{t-1}^{heat}$. Meanwhile, the current frame $t$ produces ReID loss $L_{t}^{reid}$ along with detection losses, including heatmap loss $L_{t}^{heat}$, offset loss $L_t^{off}$, and size loss $L_t^{wh}$.

Following CenterNet~\cite{zhou2019objects}, we use the L1 loss function to supervise $L_t^{off}$ and $L_t^{wh}$, and apply the focal loss function to supervise $L_{t-1}^{heat}$ and $L_{t}^{heat}$.
In terms of ReID loss, we consider each instance as a distinct class and transform the predicted feature map into a class distribution vector $\textbf{P}=\{p_i\}_i^N$ using a fully connected layer and a softmax operation. Here, $N$ is the number of all targets in the training set.
The one-hot representation of the ground-truth label can be denoted as $\textbf{Q}=\{q_i\}_i^N$, then the cross-entropy loss function is employed to supervise the ReID loss between $\textbf{P}$ and $\textbf{Q}$.
The overall loss can be expressed as:
\begin{equation}
    \begin{aligned}
        L = & \frac{1}{2}[\frac{1}{e^{\beta_1}}(L_{t-1}^{heat}+L_{t}^{heat}+L_t^{off}+L_t^{wh}) \\
            &  +\frac{1}{e^{\beta_2}}(L_{t-1}^{reid}+L_{t}^{reid})] + \beta_1 + \beta_2
    \end{aligned}
\end{equation}
where $\beta_1$ and $\beta_2$ represent learnable coefficients in~\cite{kendall2018multi} to balance different branches.

\section{Experiments}
\subsection{Implementation Details}
During training, the raw frames are resized into $3 \times 608\times1088$ pixels as inputs for the feature extractor to generate the original feature map $\textbf{FM}\in\mathbb{R}^{64 \times H \times W}$, where $H=152$ and $W=272$.
We adopt the Adam optimizer with an initial learning rate of 7e-5 to train our network. The batch size is set to 16. There are a total of 30 epochs while the learning rate decays to 7e-6 after 20 epochs. 
To obtain temporal trajectory information, we randomly sample frame pairs from the same video sequences under the condition that the interval between these sample pairs is within 3 frames.
All experiments in this work are implemented with four NVIDIA GeForce RTX 3090 GPUs.

% For our network, the video resolution is $608\times1088$, and through the backbone, the original feature map $\textbf{HM}\in\mathbb{R}^{64 \times H \times W}$ is generated by downsampling four times, where $H=152$ and $W=272$.
% First, we resize the input frames into $608\times1088$ and feed into the network to get the original feature map $\textbf{HM}\in\mathbb{R}^{64 \times H \times W}$, where $H=152$ and $W=272$. 
% We adopt the Adam optimizer with an initial learning rate of 7e-5 to train our network for 30 epochs.
% The learning rate decays to 7e-6 after 20 epochs. 
% The batch size is configured to be 16.
% To obtain temporal trajectory information during the training process, we randomly sample frame pairs from the same video sequences, and the interval between image pairs is within 3 frames.
% All experiments are implemented with four NVIDIA RTX 3090 GPUs.

\subsection{Online Inference}
In the inference phase, we follow the online data association pipeline in ByteTrack~\cite{zhang2022bytetrack}. 
Specifically, we classify the detection results into high-confidence and low-confidence detections based on a specified threshold $\tau=0.4$.
In high-confidence detection cases, we first utilize the Kalman filter to predict the location of each trajectory and compute the Mahalanobis distance between the current detections and the predicted trajectories. 
The cosine distance computed on ReID embeddings is then combined with the Mahalanobis distance to complete the initial matching. 
Subsequently, the remaining detections and trajectories are matched by intersection over union (IOU) distance.
In low-confidence detection cases, we directly employ IOU distance to associate unmatched trajectories and recover low-scoring detection boxes.
Finally, we initialize unmatched high-confidence detections as new trajectories while discarding unmatched low-confidence results.
The last remaining unmatched trajectories are stored for 30 frames.

% In the inference phase, we follow the online data association method in ByteTrack~\cite{zhang2022bytetrack}. 
% Specifically, we classify the detection results into high-confidence and low-confidence detections using a specified threshold $\tau=0.4$.
% With high-confidence detection situations, we first utilize the Kalman filter to predict the location of each trajectory in the current frame and compute the Mahalanobis distance between the current detections and the predicted trajectories.
% The cosine distance computed on ReID embeddings is combined with the Mahalanobis distance to complete the initial matching. 
% Subsequently, the remaining detections and trajectories undergo a second matching based on intersection over union (IOU) distance.
% With low-confidence detections, we directly employ IOU distance to associate unmatched trajectory, recovering low-scoring detection boxes.
% Finally, we initialize unmatched high-confidence detection results as new trajectories and eliminate unmatched low-confidence results. 
% The last remaining unmatched trajectories are stored for 30 frames.

\begin{figure*}[t]
    \centerline{\includegraphics[scale=0.53]{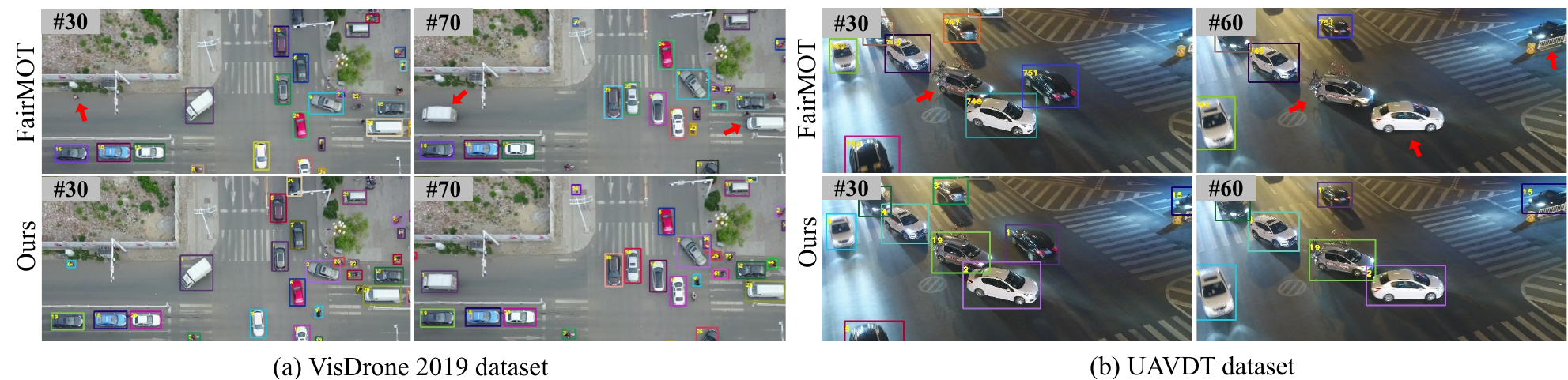}}
    \caption{Visualization of our \ourmethod and baseline FairMOT on the VisDrone2019-uav0305 set and UAVDT-M0205 sets. Each bounding box with a unique ID number represents a tracked target.}
    \label{case-vis}
    \vspace{-0.4cm}
\end{figure*}

\subsection{Comparison with State of the Arts}
We evaluate our proposed \ourmethod with current state-of-the-art MOT methods in two datasets: VisDrone2019~\cite{du2019visdrone} and UAVDT~\cite{du2018uavdt}. 
In Table~\ref{test}, multiple metrics~\cite{milan2016mot16} are presented to assess tracking performance.
Among them, MOTA and IDF1 are key metrics, which emphasize detection performance and association performance, respectively.

\noindent
\textbf{Results on VisDrone2019:}
VisDrone2019 is a fundamental benchmark for multiple object tracking based on UAV-captured videos.
It contains 10 categories, including pedestrians, vehicles, etc.
Table~\ref{test} demonstrates that \ourmethod achieves a new state-of-the-art performance with 52.0\% on IDF1 and 41.2\% on MOTA, outperforming all the previous outstanding trackers.
Concretely, our method outperforms UAVMOT~\cite{liu2022uavmot} by 1.0\% on IDF1 and 5.1\% on MOTA. Compared with OCSORT~\cite{cao2023observation}, our method improves IDF1 and MOTA by 1.6\%  and 1.6\%, respectively.
% Particularly, Compared with the UAVMOT, our method improves by 1.0\% on IDF1 and 5.1\% on MOTA. Compared with OCSORT, it surpasses by 1.6\% on IDF1 and 1.6\% on MOTA.

\noindent
\textbf{Results on UAVDT:}
The tracking videos in UAVDT are derived from bird's-eye views at different altitudes and diverse outdoor scenes. 
The tracking results of UAVDT are shown in Table~\ref{test}. 
We can observe that \ourmethod shows the best tracking performance with 69.8\% on IDF1 and 49.2\% on MOTA. 
In particular, \ourmethod improves UAVMOT~\cite{liu2022uavmot} by 2.5\% on IDF1 and 2.8\% on MOTA, while outperforming OCSORT~\cite{cao2023observation} by 4.9\% on IDF1 and 1.7\% on MOTA.

% Tracking videos in UAVDT present greater challenges, featuring bird's-eye views at different altitudes and diverse outdoor scenes.
% As shown in Table~\ref{test}, \ourmethod attains the best tracking performance, specifically 69.8\% on IDF1 and 49.2\% on MOTA. Compared with UAVMOT and OCSORT, \ourmethod outperforms them by 2.5\%/4.9\% on IDF1 and 2.8\%/1.7\% on MOTA, respectively.

\subsection{Ablation Study}

To validate the effectiveness of the individual components of our proposed tracking framework, we conduct ablation experiments on the VisDrone2019 test set.
Note that FairMOT~\cite{zhang2021fairmot} is set as our baseline method. 

% To validate the effectiveness of our proposed tracking framework, we conduct ablation experiments on the VisDrone2019 test set.
% Note that we set FairMOT~\cite{zhang2021fairmot} as our baseline method. 

\noindent
\textbf{Component ablation.}
As shown in Table~\ref{comparemodules}, when combining baseline with our proposed TEBM, the performance improves by 4.1\% on IDF1 and 4.2\% on MOTA. 
While combining baseline with the TDRM, we achieve 51.0\% on IDF1 and 38.5\% on MOTA, respectively.
% These two ablation results demonstrate their individual contribution to our network.
These showcase their respective contributions to the final \ourmethod.
Further, there is a mutually reinforcing improvement ($+5.8\%$ on IDF1 and $+5.5\%$ on MOTA) in final performance when using TEBM and TDRM together.
% Furthermore, while associated with TEBM and TDRM, the final performance gets a mutually reinforcing improvement (5.8\% on IDF1 and 5.5\% on MOTA).

\noindent
\textbf{Attributes performance ablation.}
% According to the second row of Table~\ref{comparemodules}, the proposed TEBM improves the tracking performance of the baseline by 4.1\% on IDF1 and 4.2\% on MOTA.
% Meanwhile, in the third row of Table~\ref{comparemodules}, the baseline equipped with TDRM achieves 51.0\% on IDF1 and 38.5\% on MOTA, surpassing the baseline by 4.8\% and 2.8\%, respectively.
% Such results effectively demonstrate the crucial role of exploiting temporal embedding to extract valuable features for both the detection and ReID heads.
% By inserting TEBM and TDRM into the baseline model in a cascade manner, their synergy achieves excellent tracking performance. 
% As shown in the fourth row of Table~\ref{comparemodules}, \ourmethod outperforms the baseline by 5.8\% on IDF1 and 5.5\% on MOTA. In addition, the ML, MT and FN all achieve the best performance. 
% To further understand the different aspects contribution of the proposed module on association and detection capabilities.
% We evaluate the differences in Identification Precision (IDP) and Identification Recall (IDR) corresponding to association performance, as well as precision and recall corresponding to detection performance details.
% The results of these comparisons are presented in Table~\ref{comparebase}.  
% The observation highlights that TEBM attains the top results in terms of IDP and Precision, while TDRM excels in IDR and Recall.
To investigate the contribution of the proposed modules, we evaluate the identification precision (IDP) and identification recall (IDR) w.r.t. association performance, as well as precision and recall w.r.t. detection performance.
The results of these comparisons are presented in Table~\ref{comparebase}. 
As we can see, TEBM contributes most to the IDP/Precision attributes while TDRM contributes mainly to IDR ($+7.1\%$) and Recall ($+7.5\%$).
This indicates that TEBM is primarily geared towards enhancing the accuracy of target identification, whereas TDRM exhibits a more robust capability in capturing targets.
In addition, TEBM has achieved significant improvements in all four specific metrics. 
This phenomenon shows that TEBM effectively strengthens the data association ability and demonstrates its superiority in comprehensive performance.
In contrast to the baseline, TDRM shows a lower Precision. This is mainly attributed to the trade-off between reducing the number of missed detections and an increase in false positives.
However, TDRM has significantly enhanced performance compared with the baseline in terms of both IDF1 and MOTA metrics, serving as compelling evidence of its effectiveness.
\begin{table}[t]
    \caption{Components ablation results. 
    }% \checkmark represents the baseline equipped with the proposed module.}
    \centering
    \resizebox{\columnwidth}{!}{
    \begin{tabular}{cc|ccccccc}
    \hline
      TEBM & TDRM &IDF1$\uparrow(\%)$ &MOTA$\uparrow(\%)$  &MT$\uparrow$ &ML$\downarrow$ &FP$\downarrow$ &FN$\downarrow$ &IDS$\uparrow$\\
    \hline
     \  &                       &46.2 &35.7 &509 &532 &29241 &114286 &4152  \\
     \checkmark &               &50.3(\textcolor{green}{+4.1}) &39.9(\textcolor{green}{+4.2})
     &574 &503 &\textbf{27721} &106176  &4045  \\
     \  &\checkmark             &51.0(\textcolor{green}{+4.8}) &38.5(\textcolor{green}{+2.8}) 
     &653 &475 &40113 &97105  &\textbf{3836} \\
    \checkmark & \checkmark     &\textbf{52.0}(\textcolor{green}{+5.8}) &\textbf{41.2}(\textcolor{green}{+5.5}) &\textbf{667} &\textbf{453} &36428 &\textbf{94445}  &3984\\
    \hline
    \end{tabular}}
    \label{comparemodules}
    \vspace{-0.4cm}
\end{table}
\begin{table}[t]
    \caption{The impact of components on specific association and detection metrics.}
    \centering
    \resizebox{\columnwidth}{!}{
    \begin{tabular}{c|cc|cc}
    \hline
    Module  &IDP$\uparrow(\%)$ &IDR$\uparrow(\%)$  &Precision$\uparrow(\%)$ &Recall$\uparrow(\%)$\\
    \hline
    Base      &59.8 &37.6 &79.8 &50.2\\
    Base+TEBM &\textbf{63.4}(\textcolor{green}{+3.6}) &41.7(\textcolor{green}{+4.1}) &\textbf{81.7}(\textcolor{green}{+1.9}) &53.7(\textcolor{green}{+4.1}) \\
    Base+TDRM &59.5(\textcolor{green}{-0.3}) &\textbf{44.7}(\textcolor{green}{+7.1}) &76.8(\textcolor{green}{-3.0}) &\textbf{57.7}(\textcolor{green}{+7.5})\\
    \hline
    \end{tabular}}
    \label{comparebase}
    \vspace{-0.4cm}
\end{table}

\subsection{Visualization and Analysis}
In this subsection, we qualitatively analyze the robustness of \ourmethod under challenging scenarios.
As depicted in Fig.~\ref{case-vis} (a), when faced with targets of similar appearance and small size, the FairMOT~\cite{zhang2021fairmot} fails to track them within a certain interval, while our \ourmethod can maintain the continuity of their trajectories.
The relative motion between the vehicles and the UAV platform causes target deformation and blurring, as illustrated in Fig.~\ref{case-vis} (b). FairMOT misses these vehicles, but our \ourmethod can accurately track them.
In general, FairMOT encounters difficulty in tracking targets under extreme scenarios, as indicated by the red arrows in the failure cases.
On the contrary, our method successfully tracks these targets across multiple frames and consistently preserves their unique identity numbers.
Moreover, as shown in Fig.~\ref{tsne}, compared with FairMOT, \ourmethod exhibits superior discriminative embedding representation, particularly in scenarios where targets have similar appearances.
These visualizations reveal the adaptability of our \ourmethod to complex scenarios in UAV videos, which exploits spatial and temporal information to achieve more robust and accurate tracking results.

\section{Conclusion}
In this paper, we propose a novel spatio-temporal feature cohesion framework for the UAV-based MOT task, which considers the sequence temporal consistency for both ReID and detection branches.
For the ReID branch, we introduce the temporal embedding boosting module (TEBM), which assesses the similarity among adjacent ReID feature maps to enhance the distinctiveness of individual embedding features.
For the detection branch, the temporal detection refinement module (TDRM) is proposed to propagate trajectory embedding and prominently highlight potential target positions. 
Experiments demonstrate that our proposed \ourmethod can effectively mitigate the occurrence of missed targets and improve tracking consistency.
Moreover, \ourmethod achieves new state-of-the-art performance on papular UAV-based MOT benchmarks. 
In particular, it realizes an IDF1 of 52.0\% and a MOTA of 41.2\% on the VisDrone2019 dataset, and an IDF1 of 69.3\% and a MOTA of 48.9\% on the UAVDT dataset, respectively.

{
\bibliographystyle{IEEEbib}
\bibliography{references.bib}
}

\end{document}